\pdfoutput=1

\documentclass[11pt]{article}

\usepackage[preprint]{acl}

\usepackage{times}
\usepackage{latexsym}

\usepackage[T1]{fontenc}

\usepackage[utf8]{inputenc}

\usepackage{microtype}

\usepackage{inconsolata}

\usepackage{graphicx}

\usepackage{multirow}
\usepackage{amsmath,amssymb}
\usepackage{graphicx}
\usepackage{booktabs}
\usepackage{enumitem}
\usepackage{cleveref}

\usepackage[normalem]{ulem}

\title{Imagine to Hear: Auditory Knowledge Generation can be an Effective Assistant for Language Models}

\author{
Suho Yoo$^{1}$\thanks{\hspace{1mm} equal contribution} \quad Hyunjong Ok$^{1,2\ast}$ \quad Jaeho Lee$^{2}$\thanks{\hspace{1mm} corresponding author} \\
$^1$HJ AILAB \qquad $^2$POSTECH\\
\texttt{\{uso7d0, hyunjong.ok\}@gmail.com, jaeho.lee@postech.ac.kr}
}

\begin{document}
\maketitle
\begin{abstract}
Language models pretrained on text-only corpora often struggle with tasks that require auditory commonsense knowledge.
Previous work addresses this problem by augmenting the language model to retrieve knowledge from external audio databases.
This approach has several limitations, such as the potential lack of relevant audio in databases and the high costs associated with constructing the databases.
To address these issues, we propose Imagine to Hear, a novel approach that dynamically generates auditory knowledge using generative models.
Our framework detects multiple audio-related textual spans from the given prompt and generates corresponding auditory knowledge. We develop several mechanisms to efficiently process multiple auditory knowledge, including a CLAP-based rejection sampler and a language-audio fusion module.
Our experiments show that our method achieves state-of-the-art performance on AuditoryBench without relying on external databases, highlighting the effectiveness of our generation-based approach. 
\end{abstract}
\section{Introduction}\label{sec:intro}

Can language models (LMs) understand auditory signals like humans? Recent studies suggest that the answer is negative. Although LMs do exhibit some understanding of audio signals \citep{ngo2024language}, they perform poorly in answering questions that require auditory commonsense knowledge \citep{ok2025audiobert}. For example, LMs pretrained on text-only datasets do not know which object is more likely to make a higher-pitched sound or which animal is most relevant to the given onomatopoeia. 

While such limitations can be partially addressed by augmenting language models with auditory representations retrieved from external audio database \citep{ok2025audiobert}, this approach suffers from two key limitations: (1) The database may not contain any audio that is directly relevant to the given query. (2) Constructing such databases requires a large computational cost.

To address these challenges, we present a novel approach, coined \textit{Imagine to Hear} (ITH), that leverages the imaginative capabilities of audio generative models \cite{evans2024stable, hung2024tangofluxsuperfastfaithful, liu2023audioldm2}. That is, ITH acquires auditory knowledge directly related to the given task by generating it instead of retrieving it from a database.

ITH works in three steps. First, we extract textual spans from the given input prompt that may contain auditory knowledge. Unlike visual imagination literature \citep{lu2022imagination}, we extract multiple short spans instead of a single long span so that we can generate cleaner audio without auditory interference. Next, we generate audio knowledge for each span, ensuring their relevance via CLAP-based rejection sampling \citep{elizalde2023clap}. Finally, we inject generated audio into the LM with a newly developed fusion module that can efficiently handle an arbitrary number of auditory features. 

\begin{figure*}[t]
    \centering
    \includegraphics[width=0.90\linewidth]{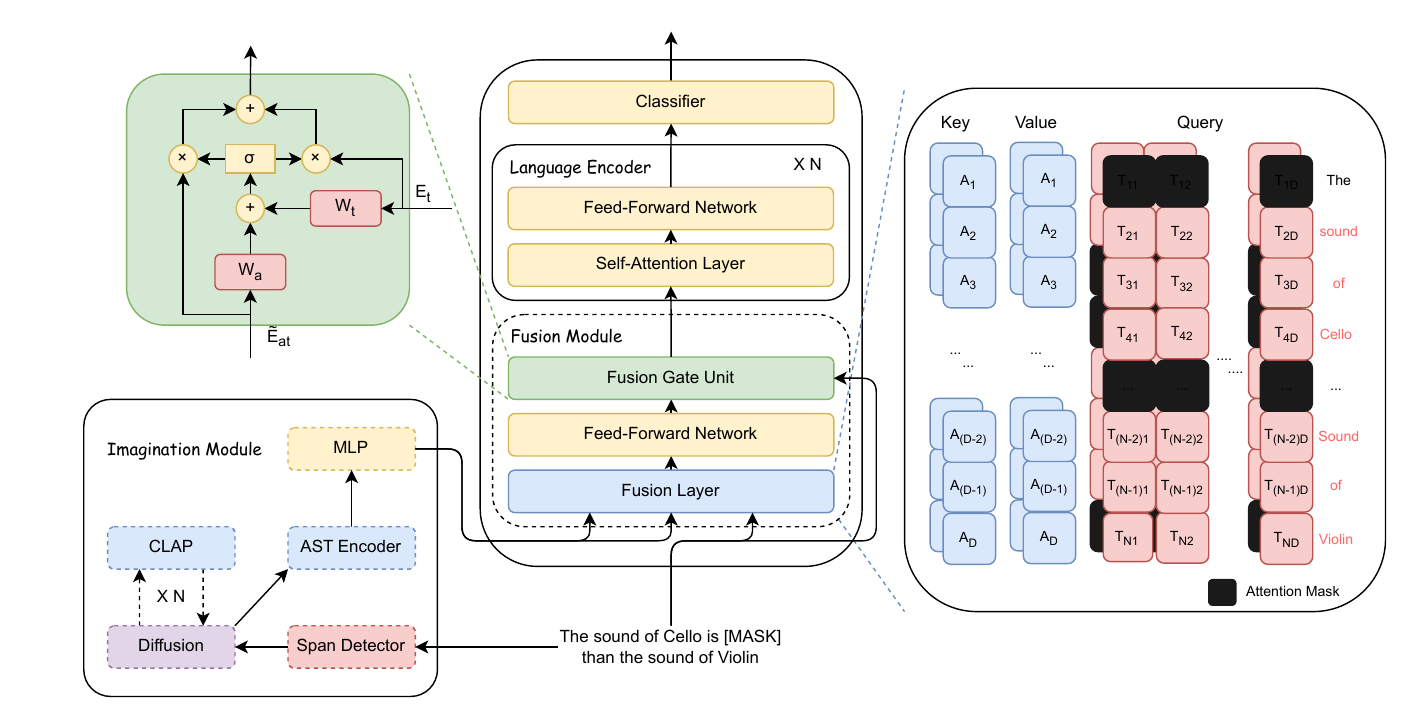}
    \caption{An illustration of the overall framework of the proposed Imagine to Hear (ITH), consisting of three components: (1) An imagination module, which detects multiple audio-related spans from the given prompt and generates multiple corresponding audio knowledge. (2) A fusion module, which combines the (variable-length) auditory and textual information. (3) A language encoder, which processes the output of the fusion module. } 
    \label{fig:model_overview}
\end{figure*}

The proposed ITH achieves state-of-the-art results on AuditoryBench \citep{ok2025audiobert}, an auditory commonsense benchmark. 
The primary contributions of our work can be summarized as follows:
\begin{itemize}[leftmargin=*,topsep=0pt,parsep=0pt,itemsep=1.5pt]
\item We develop a new generation-based framework that leverages imaginative capabilities to enhance the auditory understanding of language models.

\item We propose a new algorithm (ITH) that generates and utilizes multiple auditory knowledge from the given query, with new architectural components to process multiple audio signals while ensuring their relevance effectively.

\item ITH sets the new state-of-the-art on AuditoryBench without relying on external databases.

\end{itemize}
\section{Method: Imagine to Hear}\label{sec:method}
The proposed ITH consists of three components (\Cref{fig:model_overview}): (1) Imagination module for audio feature generation (\Cref{ssec:imag_module}); (2) Fusion module for multimodal integration (\Cref{ssec:fusion_module}); (3) Language encoder for text feature generation (\Cref{ssec:lang_encoder}). %We use pretrained LM as the language encoder---thus, we focus on describing new components: the imagination module (\Cref{ssec:imag_module}) and the fusion module (\Cref{ssec:fusion_module}).

\subsection{Imagination Module}\label{ssec:imag_module}

The imagination module (IM) works in three steps: span detection, audio generation with rejection sampling, and feature extraction.

\paragraph{Span detection.} First, IM detects audio-related spans within the textual context given as an input. Concretely, let $\mathbf{x}$ be a sequence of tokens in the textual context. We find subsequences $(\mathbf{x}^{(1)},\mathbf{x}^{(2)},\ldots)$ of $\mathbf{x}$ consisting of audio-related tokens. This is done by training a model that classifies each token of $\mathbf{x}$ as audio-related or not. Each contiguous set of audio-related tokens forms a span, and thus, we get more than one span per each textual context.

To get this classifier, we fine-tune the pretrained BERT-base \citep{devlin2019bert} on the training split of the AuditoryBench dataset.

\paragraph{Audio generation with rejection.} Next, IM uses pretrained text-to-audio diffusion models to generate audio corresponding to each auditory span. To ensure the relevance of the generated audio, we reject and resample the audio until the desired similarity to the text span has been met. The semantic similarity is measured using CLAP \citep{elizalde2023clap}. Precisely, we resample until the cosine similarity exceeds a certain threshold $\tau$, \textit{i.e.},
\begin{align}
\frac{\mathrm{CLAP}_a(\mathbf{a}_{\mathrm{gen}})^\top \mathrm{CLAP}_t(\mathbf{x}_{\mathrm{span}})}{\|\mathrm{CLAP}_a(\mathbf{a}_{\mathrm{gen}})\| \|\mathrm{CLAP}_t(\mathbf{x}_{\mathrm{span}})\|} > \tau
\end{align}
has been satisfied, where $\mathrm{CLAP}_a$ and $\mathrm{CLAP}_t$ denotes the audio and text encoders of CLAP, $\mathbf{x}_{\mathrm{span}}$ denotes the text span extracted from the input query, and $\mathbf{a}_{\mathrm{gen}}$ denotes the generated audio clip.

For efficiency, we limit the maximum number of generations; if the generative model fails to generate a well-aligned audio clip for a span in $n$ trials, we simply ignore the span. Empirically, $n=2$ has been sufficient (see \Cref{sec:ablation}).

\paragraph{Feature extraction.} The generated audio is transformed into dense embedding with an AST encoder \citep{gong21b_interspeech} followed by a two-layer MLP.

%\begin{figure}[t]
%    \centering
%    \includegraphics[width=0.95\linewidth]{figures/Imagine_to_Hear_2.pdf}
%    \caption{Overall pipeline of the Imagination Module} 
%    \label{fig:clap_rejection}
%\end{figure}

\begin{table*}[t]

\centering
\resizebox{0.85\linewidth}{!}{
\begin{tabular}{l|ccc|ccc}
\toprule
\multirow{2}{*}{\textbf{Methods}} 
& \multicolumn{3}{c|}{\textbf{Animal sound recognition}} 
& \multicolumn{3}{c}{\textbf{Sound pitch comparison}} \\

& Dev. & Test & Wiki Test 
& Dev. & Test & Wiki Test \\ 
\midrule
BERT$_{\mathrm{base}}$ \cite{devlin2019bert}     &15.51 & 13.46 & 3.05 & 59.42 & 60.41 & 48.06 \\
RoBERTa$_{\mathrm{base}}$ \cite{liu2019roberta}   
& 14.67 & 14.04 &  2.54 & 54.50 & 55.84 & 47.45 \\
Gemma2$_{2\mathrm{B}}$ \cite{team2024gemma}       
& 14.33 & 15.11 &  6.60 & 59.25 & 60.45 & 47.86 \\
LLaMA3.1$_{8\mathrm{B}}$ \cite{dubey2024llama}    
& 23.10 & 21.80 & 16.24 & 61.46 & 62.55 & 47.72 \\
AudioBERT \cite{ok2025audiobert}
& 38.28 & 36.63 & 14.32 & 73.18 & 74.83 & 55.31 \\
\midrule
Ours
& \textbf{39.36} $\pm$ 1.11 & \textbf{41.55} $\pm$ 1.06 & \textbf{19.09} $\pm$ 1.78 
& \textbf{79.82} $\pm$ 0.45 & \textbf{78.96} $\pm$ 0.48 & \textbf{76.74} $\pm$ 0.62
\\ 
\bottomrule
\end{tabular}
}
\label{tab:experiment_results_in_auditorybench}
\caption{Experimental results on AuditoryBench. The figures for baselines are taken from \citet{ok2025audiobert}.}
\end{table*}

% \begin{table}[t]
% \caption{Experiment results in AuditoryBench. The results for methods marked with $^\dag$ are from AudioBERT \cite{ok2025audiobert}.}
% \vspace{0.25cm}
% \centering
% \resizebox{\columnwidth}{!}{
% \begin{tabular}{lcccccc}
% \toprule
% \multirow{2}{*}{\textbf{Methods}} & \multicolumn{2}{|c}{\textbf{Animal sound recognition}} & \multicolumn{2}{|c}{\textbf{Sound pitch comparison}} \\ 
% & \multicolumn{1}{|c}{Test} & Wiki Test & \multicolumn{1}{|c}{Test} & Wiki Test \\ \midrule
% BERT-base$^\dag$ \cite{devlin2019bert} & 13.46 & 3.05 & 60.41 & 48.06 \\
% RoBERTa-base$^\dag$ \cite{liu2019roberta} & 14.04 & 2.54 & 55.84 & 47.45 \\
% Gemma2-2B$^\dag$ \cite{team2024gemma} & 15.11 & 6.60 & 60.45 & 47.86 \\
% LLaMA3.1-8B$^\dag$ \cite{dubey2024llama} & 21.80 & 16.24 & 62.55 & 47.72 \\
% AudioBERT$^\dag$ \cite{ok2025audiobert}& 36.63 & 14.32 & 74.83 & 55.31 \\ \midrule
% Ours & \textbf{40.74} $\pm$ 0.88 & \textbf{18.27} $\pm$ 1.28  & \textbf{78.48} $\pm$ 0.96 & \textbf{73.60} $\pm$ 1.90 \\ \bottomrule
% \end{tabular}
% }
% \label{tab:experiment_results_in_auditorybench}
% \end{table} 

\subsection{Fusion Module}\label{ssec:fusion_module}

The fusion module (FM) consists of a fusion layer, a feed-forward network, and a fusion gate unit.

\paragraph{Fusion layer.} The Fusion layer is a cross-attention module that processes textual tokens by attending to the audio tokens. In particular, the layer computes the queries of the text tokens that belong to the audio-related spans only and computes the output using only the keys and values of the audio tokens corresponding to the span.

\paragraph{Feed-forward network.} The output features from the fusion layer are processed with a two-layer MLP (same as in a typical transformer block).

\paragraph{Fusion gate unit.} The fusion gate unit combines the original input textual tokens $\mathbf{x}$ and the output of the FFN module $\mathbf{z}_{\mathrm{FFN}}$ to generate the output via gating mechanism. More concretely, the output of the module is computed as
\begin{align}
\mathbf{z}_{\mathrm{fused}} = \mathbf{g} \odot \mathbf{x} + (\mathbf{1} - \mathbf{g}) \odot \mathbf{z}_{\mathrm{FFN}},
\end{align}
where $\mathbf{g}$ is the gating signal:
\begin{align}
    \mathbf{g} = \sigma\big((\mathbf{W}_1\mathbf{x} + \mathbf{b}_1) + (\mathbf{W}_2\mathbf{z}_{\mathrm{FFN}}+\mathbf{b}_2)\big),
\end{align}
with $\sigma(\cdot)$ denoting the sigmoid activation function and $\mathbf{W}_i,\mathbf{b}_i$ denoting trainable weights and biases. This gate adaptively weighs the contributions of two modalities for each token embedding. When $\mathbf{g}$ is large, more information from the raw text will be passed to the output. When $\mathbf{g}$ is small, more information from the audio will affect the output.

\subsection{Language Encoder}\label{ssec:lang_encoder}

After the fusion module integrates the textual features with the audio features to get $\mathbf{z}_{\mathrm{fused}}$, we apply a pretrained text encoder to generate the output. Following \citet{ok2025audiobert}, we use BERT-base as our text encoder and attach a single linear layer at the end of the model as a classifier.

\subsection{Training}
The model is trained end-to-end with the training split of the AuditoryBench dataset. All weights in the model are trained, except for the audio diffusion model, CLAP, and the span detector---which has been separately trained using the AuditoryBench.

\section{Experiments}\label{sec:results}
We conducted our experiments on the AuditoryBench with further experiments. More details are provided in \Cref{sec:appendix_experiments}.
\begin{figure*}[t]
    \centering
    \includegraphics[width=1\linewidth]{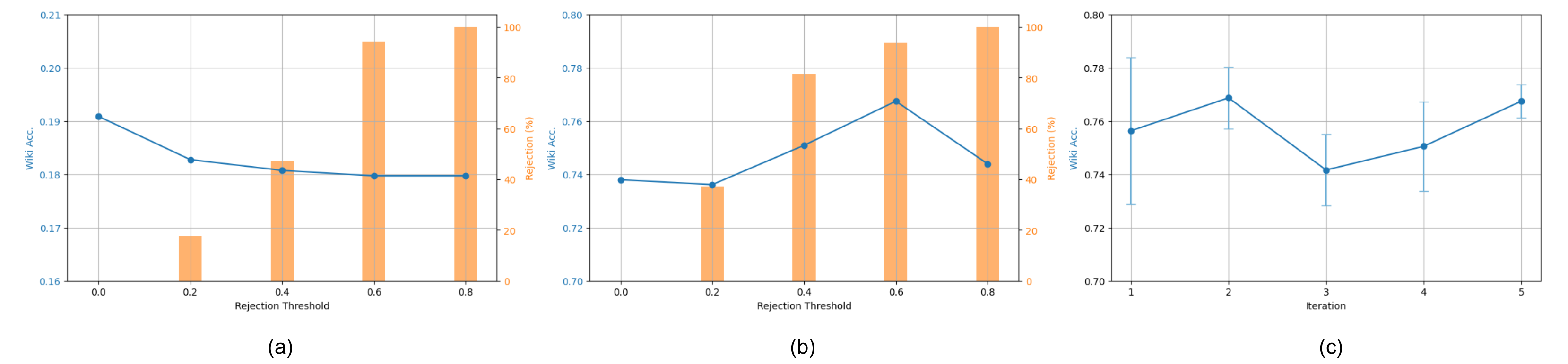}
    \caption{Ablation study of our rejection system. Each (a) and (b) are ablation results of CLAP threshold on animal sound recognition and sound pitch comparison. (c) Ablation results of iterative refinement on sound pitch comparison. The light blue line represents accuracy, while the orange bars in (a) and (b) indicate the proportion of instances where rejection occurs even after $n$ iterations, excluding the corresponding audio input.}
    \label{fig:ablation_combined}
\end{figure*}

\begin{table}[t]
\centering
\resizebox{0.9\linewidth}{!}{
\begin{tabular}{l|ccc}
\toprule
\textbf{Task} & \textbf{Dev.} & \textbf{Test} & \textbf{Wiki} \\
\midrule
Animal sound & 92.47 $\pm$ 0.77 & 92.66 $\pm$ 0.35 & 100.00 $\pm$ 0.00 \\
Sound pitch  & 94.50 $\pm$ 0.32 & 94.75 $\pm$ 0.29 & 90.24 $\pm$ 3.00 \\
\bottomrule
\end{tabular}
}
\caption{F1 scores of span detection.}
\label{tab:span_f1}
\end{table}

\paragraph{Experimental results.} 
To evaluate the effectiveness of our approach, we compare it with recent studies~\cite {ok2025audiobert} on AuditoryBench. As shown in \Cref{tab:experiment_results_in_auditorybench}, most language models perform poorly due to the absence of auditory knowledge. While AudioBERT performs well by leveraging external databases, it struggles with inter-domain, as reflected in the wiki test results. In contrast, our method leverages imagination to dynamically generate relevant auditory representations, establishing a new state-of-the-art AuditoryBench. This result highlights the enhanced generalization ability of our approach. We additionally report F1 scores of our span detector, in Table~\ref{tab:span_f1}. We observe that even a simple span detector can achieve high accuracy across all subsets.

\paragraph{Ablation studies.}\label{sec:ablation}

We first examine the effect of removing the rejection system. As shown in \Cref{fig:ablation_combined} (a), the rejection threshold of 0 is optimal for animal sound recognition. In contrast, for sound pitch comparison, the optimal threshold is 0.6, as illustrated in \Cref{fig:ablation_combined} (b). For sound pitch comparison, to further analyze the impact of the iteration count $n$, we conducted an additional ablation study by fixing the rejection threshold at 0.6 (see \Cref{fig:ablation_combined} (c)). The results indicate that accuracy improves as $n$ increases, reaching its peak at $n=2$. Detailed performance gains are reported in \Cref{tab:ablation_auditorybench}.

The fusion gate plays a crucial role in maintaining performance by selectively incorporating auditory knowledge while preventing interference.

A lack of dynamic knowledge injection generates audio for the entire sentence. In tasks like sound pitch comparison, where multiple auditory spans exist, generating audio for the entire sentence results in overlapping sounds, which significantly affects comprehension and model performance. Removing dynamic knowledge injection and the fusion gate further amplifies this issue, as indiscriminate audio injection leads to uncontrolled modality integration.

\begin{table}[t]
\centering
\resizebox{0.9\linewidth}{!}{
\begin{tabular}{l|cc|cc}
\toprule
\multirow{2}{*}{\textbf{Models}} 
& \multicolumn{2}{c|}{\textbf{Animal sound recognition}} 
& \multicolumn{2}{c}{\textbf{Sound pitch comparison}} \\

& \textbf{Test} & \textbf{Wiki Test} 
& \textbf{Test} & \textbf{Wiki Test} \\ 
\midrule
ITH (Ours)        & 41.55 $\pm$ 1.06 & 19.09 $\pm$ 1.78 & 78.96 $\pm$ 0.48 & 76.74 $\pm$ 0.62 \\
w/o Rejection     & -0.00 & -0.00 & -0.49 & -2.95    \\
w/o FG   & -0.66 & -2.14 & -0.78 & -0.87 \\
w/o DKI  & -0.84 & -0.92 & -0.13 & -1.29 \\
w/o (DKI + FG)    & -0.58 & -1.43     & -0.28 & -2.32 \\

\bottomrule
\end{tabular}
}
\label{tab:ablation_auditorybench}
\caption{The result of the ablation study. DKI denotes dynamic knowledge injection, FG denotes fusion gate. }
\end{table}

%animal clap 0
%pitch clap 0.6

\begin{figure}[t]
    \centering
    \includegraphics[width=\linewidth]{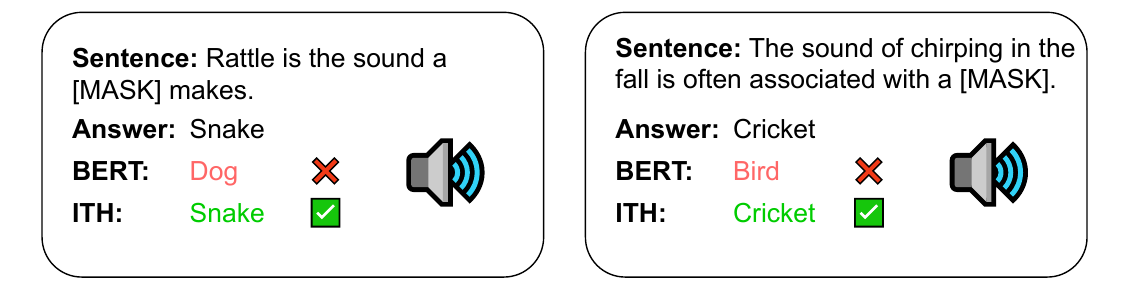}
    \caption{ITH case study. To listen to the generated sounds, visit our \href{https://imagine-to-hear.github.io}{project page}.}

    \label{fig:case_1}
\end{figure}

\begin{figure}[t]
    \centering
    \includegraphics[width=\linewidth]{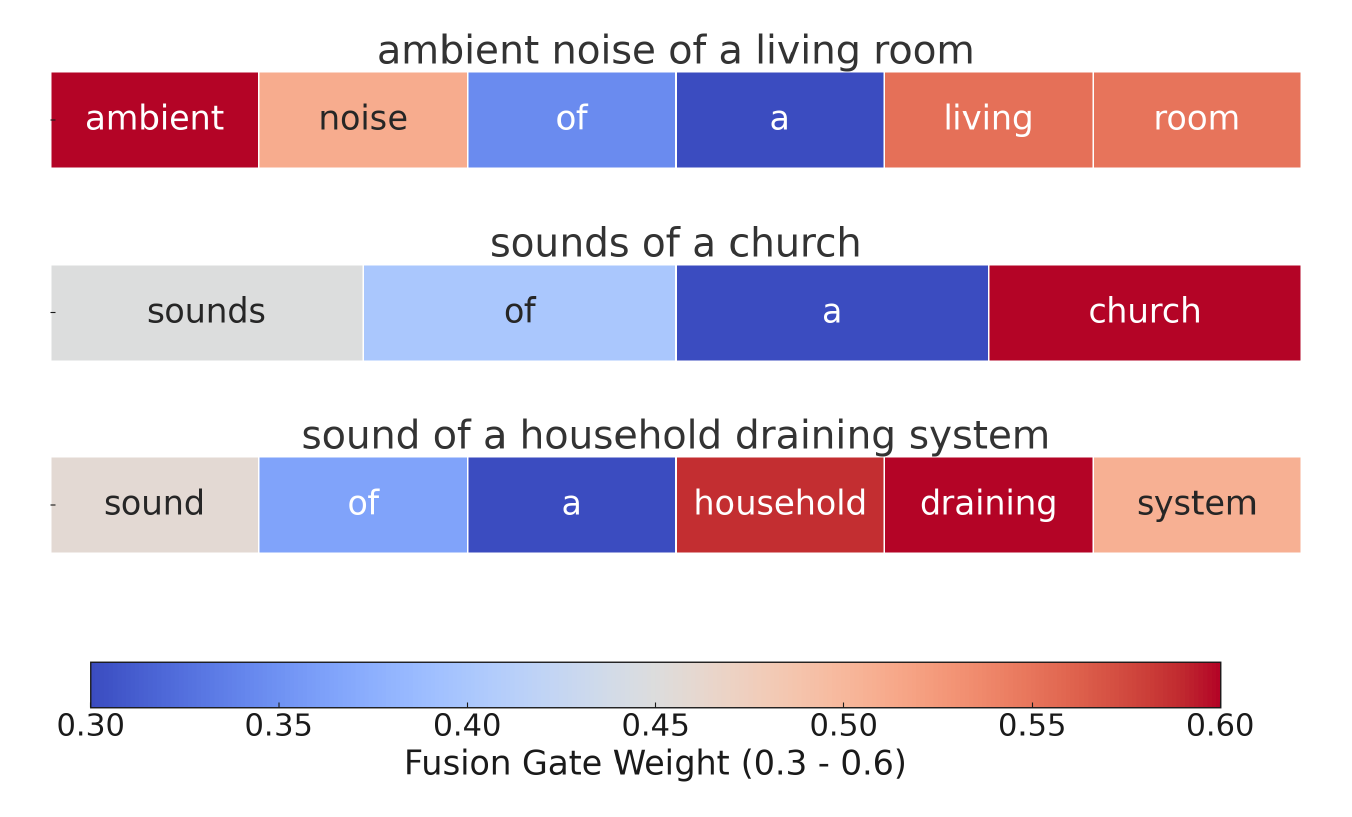}
    \caption{Fusion gate case study, where a larger fusion gate weight indicates a higher degree of audio knowledge injection into the corresponding token.}

    \label{fig:case_2}
\end{figure}

\begin{figure*}[t]
    \centering
    \includegraphics[width=\linewidth]{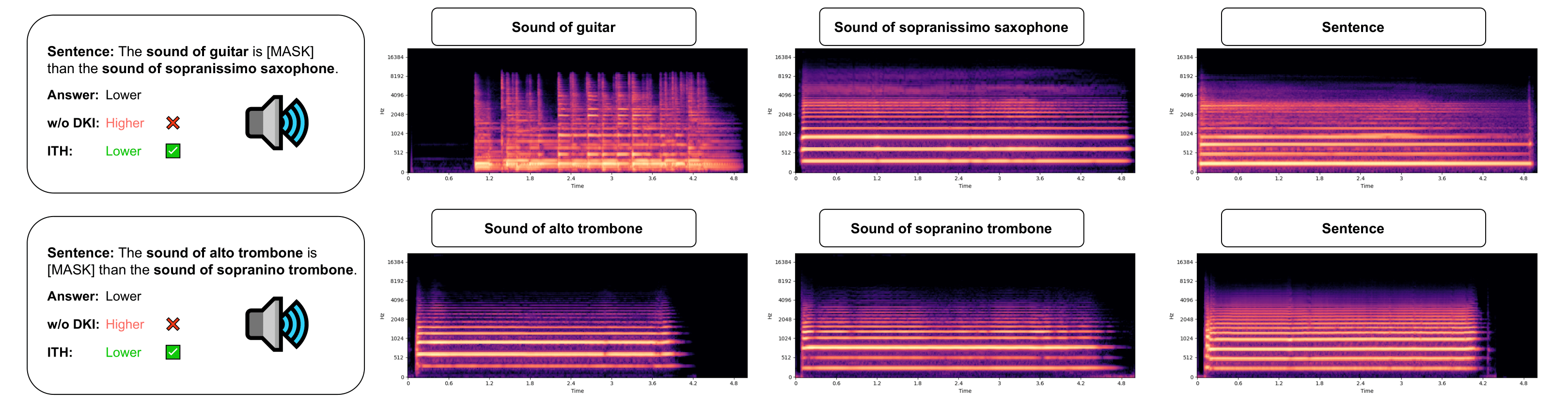}
    \caption{Dynamic knowledge injection (DKI) case study. To listen to the generated sounds, visit our \href{https://imagine-to-hear.github.io}{project page}.}
    \label{fig:case_3}
\end{figure*}

\paragraph{Case study.}
We conducted a case study to evaluate the effectiveness of our method.
As shown in \Cref{fig:case_1}, the first example describes the sound of a "rattle," strongly associated with a snake. Without auditory information, it can be challenging to make the correct prediction. However, leveraging imagined auditory knowledge, our method correctly identifies "snake" as the answer. In the second example, the sentence mentions "chirping in the fall," a sound commonly associated with crickets rather than birds. While "chirping" might semantically align with avian species, our model, by incorporating auditory reasoning, correctly predicts "cricket," demonstrating its ability to refine predictions based on imagined auditory context.

\paragraph*{Role of fusion gate.}
To analyze the role of the fusion gate in cross-modal integration, we visualize the token-wise fusion gate weights in \Cref{fig:case_2}. The fusion gate operates at the embedding level. However, for visualization purposes, we compute the average fusion gate values per token and represent them in a heatmap. 

Our analysis shows that the fusion gate effectively prioritizes auditory-relevant tokens while minimizing the influence of function words. In phrases like "ambient noise of a living room" and "sound of a household draining system," key auditory descriptors such as "ambient," "living," "room," "household," "draining," and "system" receive the highest weights. This indicates that the model assigns more importance to tokens that directly describe sound characteristics while suppressing function words like "of" and "a," which contribute less to auditory integration. 

\paragraph{Dynamic knowledge injection.}
To deepen our understanding on the role of dynamic knowledge injection (DKI), we analyze cases where injecting auditory knowledge at the span level improves model predictions, as illustrated in \Cref{fig:case_3}.

In the first example, the sentence compares the sound of a guitar with that of a sopranissimo saxophone. When generating audio representations separately for each instrument, their mel-spectrograms exhibit distinct frequency characteristics, with the saxophone producing significantly higher frequencies than the guitar. However, when generating a single audio representation for the entire sentence, this distinction is lost, leading to an incorrect prediction of ``higher.'' In contrast, our method, by dynamically injecting span-specific auditory knowledge, correctly predicts ``lower,'' maintaining the frequency contrast observed in the spectrograms.

In the second case, the sentence contrasts the sound of an alto trombone with that of a sopranino trombone. Similarly, generating audio separately for each instrument produces clear frequency differences, with the sopranino trombone having higher frequencies. However, sentence-level generation fails to capture this difference, leading to an incorrect prediction of ``higher.'' By selectively injecting auditory spans, our model correctly predicts ``lower,'' aligning with the frequency distribution observed in the mel-spectrogram analysis.

\paragraph{More details.} We also illustrate more ablation studies for each method to demonstrate their contributions in \Cref{ablation_rest}.

\section{Conclusion}\label{sec:conclusion}
In this work, we proposed \textit{Imagine to Hear}, a novel generation-based approach for injecting auditory knowledge into language models. By dynamically generating audio, our method overcomes the limitations of retrieval-based systems. With a hierarchical architecture and a rejection system, we ensure high-quality, context-relevant audio knowledge integration. Experiments demonstrate significant improvements on AuditoryBench with audio imagination. This work underscores the value of imagination in enhancing multimodal capabilities in language models. 

ITH extends beyond its established advantages over retrieval methods in the multimodal era. Its utility is notable in resource-constrained settings reliant on unimodal models where multimodal inputs are infeasible. Furthermore, even for advanced multimodal systems, ITH can provide a practical pathway when audio is absent, ambiguous, or subtle—parallel to visual imagination modules that enhance reasoning without direct visual input \citep{lu2022imagination, zhu2023visualize, liu2024enhancingvisualreasoningautonomous}.We expect that our approach will remain effective and complementary in the era of multimodal language models.

\section*{Limitations}
One limitation is that the effectiveness of our method is influenced by the quality of input prompts. If the input lacks sufficient auditory cues, the generated audio may not always align perfectly with the intended knowledge (\Cref{sec:appendix_glue}). Additionally, while our rejection-based generation ensures high-quality auditory integration, it introduces some computational overhead.

% \section*{Ethics statement}

\section*{Acknowledgments}
This work was supported in part by the Institute of Information \& Communications Technology Planning \& Evaluation (IITP) grant funded by the Ministry of Science and ICT (MSIT)(No.RS-2019-II191906, Artificial Intelligence Graduate School Program (POSTECH)), and in part by the National Research Foundation of Korea (NRF) grant funded by the Korea government (MSIT)(No. RS-2023-00213710, neural network optimization with minimal optimization costs). We also thank our colleagues at EffL Lab and others for their help.

\bibliography{custom}

% \newpage
% \clearpage

\appendix

\section{Related work}\label{sec:related}

\paragraph{Auditory knowledge in language model.}
Pretrained language models (LMs) lack auditory commonsense knowledge due to their text-only training, limiting their ability to reason about sound-related concepts. Recent studies have investigated whether LMs encode latent auditory information by aligning text embeddings with representations from pretrained audio models. While these studies suggest some degree of alignment, the learned representations remain insufficient for effectively understanding and utilizing auditory information \cite{ngo2024language}.

Retrieval-based approaches attempt to address this limitation by retrieving relevant audio-text pairs from external databases and integrating them into LMs. AudioBERT \cite{ok2025audiobert} follows this paradigm, detecting spans that require auditory knowledge and injecting retrieved information through lightweight adaptation layers. However, these methods inherently depend on the availability of existing data, making them unreliable when relevant audio is missing.

Our approach instead generates auditory knowledge directly from context using a text-to-audio diffusion model, removing the need for external retrieval. By dynamically incorporating auditory information, this method offers a generalizable solution that can be applied across various tasks requiring auditory understanding.

\paragraph{Visual knowledge in language model.}
Beyond auditory knowledge, LMs also lack visual commonsense knowledge \cite{zhang2022visual, liu2022things, alper2023bert, rahmanzadehgervi2024vision}, which limits their ability to reason about concepts such as object shapes, colors, and spatial relationships. For this problem, previous studies have explored methods for integrating visual knowledge, primarily by generating images from textual input and incorporating them into LMs.

Most existing approaches generate images based on the entire input context or sentence level generation, assuming that all textual content requires visual grounding. While this allows for broad visual augmentation, it often results in irrelevant or redundant images, as these methods do not selectively identify which parts of the text actually require visual support \cite{wang2023visually, tang2023learning_imagine, lu2022imagination}. A different line of work detects visually informative spans within the text before aligning them with visual representations \cite{guo2023visually}, but these methods lack mechanisms to verify the relevance of generated images, leading to potential inconsistencies when integrating visual knowledge.

Our approach addresses these limitations by detecting specific spans and generating representation only for those spans. Additionally, we introduce a rejection system that filters out irrelevant representation before integration, ensuring that only contextually aligned generated knowledge is incorporated into the model.

\paragraph{Generation models.}
Diffusion models have become the dominant paradigm for generative tasks, yet recent studies have increasingly focused on improving efficiency in training and inference. In text-to-audio generation, models like AudioLDM2 \cite{liu2023audioldm2}, Tango \cite{hung2024tangofluxsuperfastfaithful}, and Stable Audio \cite{evans2024stable} optimize stable diffusion-based architectures, leveraging latent representations to reduce computational overhead while maintaining high-quality generation.

\section{More ablation study}
\label{ablation_rest}

\begin{table}[t]
\centering
\resizebox{\linewidth}{!}{
\begin{tabular}{l|cc|cc}
\toprule
\multirow{2}{*}{\textbf{Models}} 
& \multicolumn{2}{c|}{\textbf{Animal sound recognition}} 
& \multicolumn{2}{c}{\textbf{Sound pitch comparison}} \\

& \textbf{Test} & \textbf{Wiki Test} 
& \textbf{Test} & \textbf{Wiki Test} \\ 
\midrule
Stable       & \textbf{41.55} $\pm$ 1.06 & \textbf{19.09} $\pm$ 1.78 & \textbf{78.47} $\pm$ 0.44 & 73.79 $\pm$ 0.99  \\
AudioLDM 2    & 40.89 $\pm$ 0.57 &  18.38 $\pm$ 2.77 & 78.44 $\pm$ 0.90 & 73.42 $\pm$ 1.05  \\
MusicGen       & 41.01 $\pm$ 1.27 &  17.77 $\pm$ 1.19 & 78.05 $\pm$ 0.70 & \textbf{74.06} $\pm$ 1.27 \\
\bottomrule
\end{tabular}
}
\label{tab:ablation_diffusion}
\caption{diffusion ablation}
\end{table}

% clap 0.0

\paragraph{Various generation models.}

To evaluate the impact of different audio generation models, we conducted an ablation study comparing Stable Audio \cite{evans2024stable}, AudioLDM 2 \citep{liu2023audioldm2}, and MusicGen \cite{copet2024simplecontrollablemusicgeneration} in \Cref{tab:ablation_diffusion}. 

Overall, Stable achieved the best performance across most tasks, demonstrating its effectiveness in generating relevant auditory representations. It outperformed other models in both animal sound recognition and sound pitch comparison, particularly in the Test sets. This suggests that Stable provides the most robust auditory features, contributing to better downstream task performance.

Interestingly, MusicGen performed slightly better than other models in the sound pitch comparison wiki test. This dataset primarily involves musical instrument pitch comparison, aligning well with MusicGen's domain specialization in music generation. The results indicate that MusicGen can generate more domain-relevant auditory representations for musical contexts, leading to an improvement in this specific evaluation.

\section{Human Evaluation}

\paragraph*{Generated Audio.}
To assess perceptual quality of the generated audio samples, we have conducted human evaluations comparing our generated audio with real audio retrieved from LAION-Audio-630K. Sixteen participants have evaluated the samples under both in pairwise and 3-way identification settings, with the total of 40 questions. The results, shown in Table~\ref{tab:pairwise} and Table~\ref{tab:threeway}, indicate that the generated audio is often indistinguishable from the real recordings. Specifically, Table~\ref{tab:pairwise} shows that, when comparing retrieval-based audio (in AudioBERT) and generated audio (in ours) conditioned on the same input text from AuditoryBench, human evaluators exhibit a clear preference for the generated audio. Table~\ref{tab:threeway} presents a 3-way identification task where participants are asked to identify the real sample among two generated and one real audio; the results suggest that participants struggle to distinguish real from generated audio reliably.

\begin{table}[t]
\centering
\resizebox{0.5\linewidth}{!}{
\begin{tabular}{l|c}
\toprule
\textbf{Preference} & \textbf{Ratio (\%)} \\
\midrule
Real Audio      & 41.16 \\
Generated Audio & 50.30 \\
Equal     & 8.54  \\
\bottomrule
\end{tabular}
}
\caption{Pairwise comparison between real and generated audio.}
\label{tab:pairwise}
\end{table}

\begin{table}[t]

\centering
\resizebox{\linewidth}{!}{
\begin{tabular}{l|cc}
\toprule
\textbf{Setup} & \textbf{Chose Real (\%)} & \textbf{Chose Generated (\%)} \\
\midrule
2 Reals, 1 Gen & 65.00    & \textbf{35.00} \\
2 Gens, 1 Real & \textbf{44.03} & 55.97    \\
\bottomrule
\end{tabular}
}
\caption{We design a 3-way identification task to evaluate the realism of generated audio. In the first setup, participants are presented with two real audio samples and one generated sample and are asked to identify which one is the generated audio. In the second setup, the inverse task is given: among the three audio samples (two generated and one real), participants are asked to identify the real audio.}
\label{tab:threeway}
\end{table}

We have further evaluated on the wiki set, which includes real audio aligned with the prompts. As shown in Table~\ref{tab:wiki_eval}, ten out of sixteen annotators have judged the generated audio to be of sufficiently high quality and fidelity to serve as plausible substitutes; we have tested with 20 questions sampled from the wiki set. These findings confirm that our method produces realistic and contextually appropriate auditory knowledge.

\begin{table}[t]
\centering
\resizebox{0.8\linewidth}{!}{
\begin{tabular}{l|c}
\toprule
\textbf{Evaluation Criterion} & \textbf{Result} \\
\midrule
Acceptable as substitute for real audio & 77.42\% \\
Quality and semantic fidelity (1--5 scale) & 3.78 \\
\bottomrule
\end{tabular}
}
\caption{Human evaluation on the Wikipedia subset of AuditoryBench.}
\label{tab:wiki_eval}
\end{table}

\begin{table*}[t]
\centering
\vspace{0.25cm}
\resizebox{\linewidth}{!}{
\begin{tabular}{l|ccccc|ccccc}
\toprule
\textbf{Category} & \textbf{Train} & \textbf{Dev} & \textbf{Test} & \textbf{Wiki} & \textbf{Total} & \textbf{Train} & \textbf{Dev} & \textbf{Test} & \textbf{Wiki} & \textbf{Total} \\ \midrule
\multicolumn{1}{l}{} & \multicolumn{5}{|c|}{\textbf{Animal sound recognition}} & \multicolumn{5}{c}{\textbf{Sound pitch comparison}}\\ \midrule
\# Sentence & 4,211 & 593 & 1,211 & 197 & 6,212 & 8,312 & 1,178 & 2,387 & 3,625 & 15,502  \\ 
\# Words/Sentence & 9.27 & 9.39 & 9.30 & 7.01 & 9.21 & 18.18 & 18.12 & 18.19 & 12.41 & 16.83 \\ 
\# Total words & 39,024 & 5,567 & 11,268 & 1,381 & 57,240 & 151,081 & 21,343 & 43,431 & 44,987 & 260,842  \\ 
\bottomrule
\end{tabular}
}
\label{tab:animal_stats_combined}
\caption{Statistics for Auditorybench.}
\end{table*}

\begin{table*}[t]
\centering
\setlength{\tabcolsep}{5pt}
\scalebox{0.9}{
\begin{tabular}{l | c c c c c c c}
\toprule
\textbf{Methods} & \textbf{SST-2} & \textbf{QNLI} & \textbf{QQP} & \textbf{MNLI} & \textbf{MRPC} & \textbf{STS-B} & \textbf{Avg.}\\
\midrule 
\multicolumn{8}{c}{Multimodal text encoder}\\
\midrule
CLIP \cite{radford2021clip} & 73.3 & 74.5 & 72.8 & 68.4 & 74.3 & 73.8 & 72.85 \\
BLIP \cite{li2022blip}      & 76.3 & 77.4 & 78.8 & 72.5 & 77.8 & 76.4 & 76.53 \\
ALBEF$_{14\mathrm{M}}$ \cite{ALBEF}
                           & 78.9 & 78.2 & 79.4 & 73.4 & 76.5 & 77.5 & 77.31 \\
\midrule 
\multicolumn{8}{c}{BERT$_{\mathrm{base}}$}\\
\midrule
Baseline \cite{devlin2019bert} & 89.3 & 87.9 & 87.2 & 79.4 & 81.7 & 84.4 & 84.98 \\
VOKEN \cite{tan2020vokenization} & 92.2 & 88.6 & 88.6 & 82.6 & 83.5 & 86.0 & 86.83 \\
iACE \cite{lu2022imagination} & 91.7 & 88.6 & 89.1 & 82.8 & \textbf{85.8} & 86.6 & 87.43 \\
VAWI \cite{guo2023visually} & \textbf{92.9} & 89.1 & 89.7 & 83.0 & \textbf{85.8} & 87.2 & 87.80 \\
Ours & 92.7 & \textbf{90.4} & \textbf{90.9} & \textbf{83.7} & 85.7 & \textbf{87.5} & \textbf{88.5} \\
%Ours (+ visual) & - & 90.45 & 90.92 & - & 86.5 & 87.56 & - \\
\midrule 
\multicolumn{8}{c}{RoBERTa$_{\mathrm{base}}$}\\
\midrule
Baseline \cite{liu2019roberta} & 89.2 & 87.5 & 86.2 & 79.0 & 81.4 & 85.4 & 84.78 \\
VOKEN \cite{tan2020vokenization} & 90.5 & 88.2 & 87.8 & 81.0 & 87.0 & 86.9 & 87.83 \\
iACE \cite{lu2022imagination} & 91.6 & 89.1 & 87.9 & 82.6 & 87.7 & 86.9 & 87.63 \\
VAWI \cite{guo2023visually} & 91.7 & 90.6 & 87.9 & 82.6 & \textbf{88.5} & 88.3 & 88.21 \\
Ours & \textbf{94.6} & \textbf{91.9} & \textbf{91.4} & \textbf{87.4} & 88.0 & \textbf{90.1} & \textbf{90.6} \\
%Ours (+ visual) & - & 92.11 & 91.29 & - & 89.02 & 89.70 & - \\
\bottomrule
\end{tabular}
}
\caption{Experimental results on GLUE. The results of prior work are reported from VAWI \cite{guo2023visually}. VAWI utilizes various methods; we report the best performance of VAWI for each task.} 
\label{tab:glue_results}
\end{table*}

% \begin{table*}[t]
% \centering
% \caption{Experiment results in GLUE. The results for methods marked with $^\dag$ are from VAWI \cite{guo2023visually}. VAWI utilizes various methods; we report the best performance of VAWI for each task.} 
% \setlength{\tabcolsep}{5pt}
% \scalebox{0.92}{
% \begin{tabular}{l l c c c c c c c}
% \toprule
% \textbf{Methods} & \textbf{SST-2} & \textbf{QNLI} & \textbf{QQP} & \textbf{MNLI} & \textbf{MRPC} & \textbf{STS-B} & \textbf{Avg.}\\
% \midrule
% CLIP$^\dag$                     & 73.3 & 74.5 & 72.8 & 68.4 & 74.3 & 73.8 & 72.85 \\
% BLIP$^\dag$                     & 76.3 & 77.4 & 78.8 & 72.5 & 77.8 & 76.4 & 76.53 \\
% ALBEF$_{14\mathrm{M}}^\dag$     & 78.9 & 78.2 & 79.4 & 73.4 & 76.5 & 77.5 & 77.31 \\
% %\midrule
% BERT-base$^\dag$ & 89.3 & 87.9 & 87.2 & 79.4 & 81.7 & 84.4 & 84.98 \\
% VOKEN$^\dag$                    & 92.2 & 88.6 & 88.6 & 82.6 & 83.5 & 86.0 & 86.83 \\
% iACE$^\dag$                     & 91.7 & 88.6 & 89.1 & 82.8 & 85.8 & 86.6 & 87.43 \\
% VAWI$^\dag$                     & 92.9 & 89.1 & 89.7 & 83.0 & 85.8 & 87.2 & 87.80 \\
% \midrule
% Ours                            & 92.4 & 89.1 & 89.7 & 83.0 & 85.6 & 86.9 & 87.78 \\
% Ours +visual                    & 92.4 & 89.1 & 89.7 & 83.0 & 85.6 & 86.9 & 87.78 \\
% \bottomrule
% \end{tabular}
% }
% \label{tab:glue_results}
% \end{table*}

\begin{table*}[t]
\centering
\resizebox{\textwidth}{!}{
\begin{tabular}{l|ccc|ccc|ccc}
\toprule
\textbf{Category} & \multicolumn{3}{c|}{\textbf{SST-2}} & \multicolumn{3}{c|}{\textbf{QNLI}} & \multicolumn{3}{c}{\textbf{QQP}} \\
\midrule
 & \textbf{Train} & \textbf{Dev} & \textbf{Test} & \textbf{Train} & \textbf{Dev} & \textbf{Test} & \textbf{Train} & \textbf{Dev} & \textbf{Test} \\
\midrule
\# Total Words & 633,724 & 17,046 & 35,025 & 3,818,145 & 205,581 & 204,685 & 8,050,562 & 893,635 & 8,752,112 \\
\# Span Words Ratio & 4.32 & 2.29 & 2.70 & 0.84 & 0.19 & 0.20 & 0.45 & 0.34 & 0.39 \\
\# Total Span Words & 27,378 & 391 & 947 & 31,976 & 394 & 413 & 36,068 & 3,034 & 34,353 \\
\# Total Span Count & 9,612 & 173 & 413 & 8,343 & 169 & 175 & 10,961 & 1,045 & 12,187 \\
\midrule
\textbf{Category} & \multicolumn{3}{c|}{\textbf{MNLI}} & \multicolumn{3}{c|}{\textbf{MRPC}} & \multicolumn{3}{c}{\textbf{STS-B}} \\
\midrule
 & \textbf{Train} & \textbf{Dev} & \textbf{Test} & \textbf{Train} & \textbf{Dev} & \textbf{Test} & \textbf{Train} & \textbf{Dev} & \textbf{Test} \\
\midrule
\# Total Words & 11,695,881 & 286,426 & 289,461 & 160,986 & 17,968 & 75,042 & 114,346 & 34,147 & 27,052 \\
\# Span Words Ratio & 2.25 & 1.63 & 1.60 & 0.94 & 0.55 & 0.47 & 2.73 & 2.31 & 2.61 \\
\# Total Span Words & 263,536 & 4,662 & 4,633 & 1,521 & 98 & 354 & 3,124 & 789 & 706 \\
\# Total Span Count & 72,473 & 1,621 & 1,577 & 582 & 63 & 231 & 1,169 & 337 & 302 \\
\bottomrule
\end{tabular}
}
\label{tab:glue_stats_with_spans}
\caption{Statistics for GLUE benchmark datasets with audio span information.}
\end{table*}

\begin{table*}[t]
\centering
\vspace{0.25cm}
\resizebox{\textwidth}{!}{ % 표 크기 조절
\begin{tabular}{l|ccc|ccc|ccc}
\toprule
\textbf{Category} & \multicolumn{3}{c|}{\textbf{SST-2}} & \multicolumn{3}{c|}{\textbf{QNLI}} & \multicolumn{3}{c}{\textbf{QQP}} \\ 
\midrule
 & \textbf{Train} & \textbf{Dev} & \textbf{Test} & \textbf{Train} & \textbf{Dev} & \textbf{Test} & \textbf{Train} & \textbf{Dev} & \textbf{Test} \\ \midrule
\# Sentence & 67,349 & 872 & 1,821 & 209,486 & 10,926 & 10,926 & 727,692 & 80,860 & 781,930 \\ 
\# Words/Sentence & 9.41 & 9.36 & 9.23 & 18.22 & 18.81 & 18.74 & 11.06 & 11.06 & 11.19 \\ 
\# Total words & 633,724 & 17,046 & 35,025 & 3,818,145 & 205,581 & 204,685 & 8,050,562 & 893,635 & 8,752,112 \\ 
\midrule
\textbf{Category} & \multicolumn{3}{c|}{\textbf{MNLI}} & \multicolumn{3}{c|}{\textbf{MRPC}} & \multicolumn{3}{c}{\textbf{STS-B}} \\ 
\midrule
 & \textbf{Train} & \textbf{Dev} & \textbf{Test} & \textbf{Train} & \textbf{Dev} & \textbf{Test} & \textbf{Train} & \textbf{Dev} & \textbf{Test} \\ \midrule
\# Sentence & 785,364 & 19,630 & 19,592 & 7,336 & 816 & 3,450 & 11,498 & 3,000 & 2,758 \\ 
\# Words/Sentence & 14.89 & 14.59 & 14.78 & 21.94 & 22.02 & 21.76 & 9.94 & 11.38 & 9.80 \\ 
\# Total words & 11,695,881 & 286,426 & 289,461 & 160,986 & 17,968 & 75,042 & 114,346 & 34,147 & 27,052 \\ 
\bottomrule
\end{tabular}
}
\label{tab:glue_stats}
\caption{Statistics for GLUE benchmark datasets.}
\end{table*}

\section{Experiment details}\label{sec:appendix_experiments}

\subsection{Training Datasets}

\paragraph*{AuditoryBench.}
We conducted our experiments on the AuditoryBench—a benchmark designed to evaluate the auditory knowledge of language models which comprises two fundamental tasks: Animal Sound Recognition, which assesses the model's ability to classify various animal sounds, and Sound Pitch Comparison, which measures its capability to discern differences in pitch between sounds. To evaluate our proposed methods, we employ accuracy for AuditoryBench. 

The dataset includes 6,212 sentences for Animal Sound Recognition and 15,502 sentences for Sound Pitch Comparison, with varying lengths and complexities. The average number of words per sentence is 9.21 for Animal Sound Recognition and 16.83 for Sound Pitch Comparison. More statistics info is in \Cref{tab:animal_stats_combined}.

\paragraph*{Implementation details.} 
The audio span detector employ a BERT-base model, trained for \(5\) epochs with a batch size of \(16\), a learning rate of \(1 \times 10^{-5}\), and using the AdamW \cite{loshchilov2018decoupled}  optimizer. For labeling, spans in the AuditoryBench dataset were directly utilized as they are included in the dataset. 

For audio generation, we employed the stable-audio-open-1.0 \cite{evans2024stable}. Compared to existing state-of-the-art models, this model was selected based on their strong performance demonstrated through experiments on various datasets. An AST \cite{gong21b_interspeech} was employed to inject auditory representations into the language model.

For training, we utilized a BERT-base model to ensure a fair comparison with existing methods \cite{ok2025audiobert}. The model was trained for \(8\) epochs with a batch size of \(32\), a learning rate of \(3 \times 10^{-4}\) and \(4 \times 10^{-5}\) for each task, and using the AdamW optimizer. 

All experiments report the average score from 5 runs with different random seeds for each setting and experiments were done on a single NVIDIA L40S or NVIDIA 6000ADA GPU or Geforce RTX 4090. Additionally, we performed a search over the CLAP rejection threshold, as well as the number of iterations, to identify the optimal values.

\section{Experiments on GLUE}
\label{sec:appendix_glue}
\paragraph{Experimental result}
To assess the impact of auditory knowledge in standard NLP tasks, we evaluated our method on the GLUE (\Cref{tab:glue_results}). While our approach shows improvements over the baseline, it is not entirely clear whether the performance gain is directly attributable to the integration of auditory knowledge. 

Indeed, audio spans appear in only 0.19\% to 4.32\% of all training and development samples in GLUE (see \Cref{tab:glue_stats_with_spans} for details), meaning that a large portion of the dataset lacks explicit auditory information. This raises the possibility that the observed improvements could stem from factors unrelated to auditory augmentation, such as increased model complexity or additional fine-tuning effects.

\paragraph{Implementation details.}
Following recent works \citep{wang2018glue, guo2023visually}, we train with a batch size of 32 and apply a weight decay of 0.01.  
A grid search is conducted to determine the optimal learning rate within the range of \([2 \times 10^{-5}, 5 \times 10^{-5}]\), and we set the number of epochs to 10.  

To detect auditory spans within GLUE, we train an audio span detector based on a BERT-base model.  
The detector is trained for 5 epochs with a batch size of 16, a learning rate of \(1 \times 10^{-5}\), and the AdamW optimizer \cite{loshchilov2018decoupled}.  
Since GLUE lacks predefined auditory span labels, we generate labels using Qwen2-72B-Instruct-AWQ \cite{yang2024qwen2} that can be inferred on a single A100 GPU. The model independently generates labels for auditory spans using a predefined prompt, which is detailed in \Cref{detail generate prompt}.

\section{Detailed dataset information of GLUE}
\label{detail data information}

The General Language Understanding Evaluation (GLUE) benchmark is utilized to assess the natural language understanding (NLU) capabilities of our model. We selected six tasks for evaluation, including SST-2, QNLI, QQP, MNLI, MRPC, and STS-B.

SST-2 (Stanford Sentiment Treebank): A binary classification task that involves determining the sentiment (positive or negative) of a given sentence extracted from movie reviews.

QNLI (Question Natural Language Inference): A task where the model predicts whether the context sentence contains the answer to a given question, derived from the Stanford Question Answering Dataset (SQuAD).

QQP (Quora Question Pairs): This task focuses on identifying whether a pair of Quora questions are semantically equivalent.

MNLI (Multi-Genre Natural Language Inference): A large-scale dataset for evaluating the model's ability to perform textual entailment across multiple genres, determining whether a given hypothesis is true, false, or undetermined based on a premise.

MRPC (Microsoft Research Paraphrase Corpus): A binary classification task where the model determines whether two given sentences are semantically equivalent.

STS-B (Semantic Textual Similarity Benchmark): A regression task that measures the degree of semantic similarity between two sentences on a scale from 0 to 5.

Evaluation Metrics: For the classification tasks (SST-2, QNLI, QQP, MNLI, MRPC), we utilize accuracy as the primary evaluation metric. For the regression task (STS-B), we employ the Spearman correlation coefficient to evaluate the correlation between the predicted similarity scores and the ground truth.

More details, including dataset statistics, are presented in \Cref{tab:glue_stats}.  
The table provides an overview of total word counts, auditory span ratios, and detected span counts for each task.

\begin{table*}[h]
\centering
\begin{tabular}{p{0.95\textwidth}}  % 한 줄(컬럼)짜리 tabular
\toprule
\textbf{Prompt for auditory knowledge}\\ 
\midrule

You will be given information about specific sentences.

Your task is to extract audio-related spans from the given sentences.

Please make sure you read and understand these instructions carefully. Please keep this document open while reviewing, and refer to it as needed.

\textbf{Extraction Criteria}
\begin{enumerate}
    \item Identify spans within the sentence that describe or represent audio-related information.
    \item The spans must directly reference sounds, audio characteristics, or audible phenomena.
\end{enumerate}

\textbf{Extraction Steps}
\begin{enumerate}
    \item Read the provided sentence.
    \item Identify the portion of the sentence that directly refers to audio-related information.
    \item If no audio-related spans are found, return \textbf{Span: None}.
    \item If audio-related spans are found, extract and format them as \textbf{Span: \{\{audio-related span\}\}}.
\end{enumerate}

\textbf{Example}

\textbf{Example 1:} \\
Sentence: A dog makes a growling bowwow sound. \\
Span: \texttt{growling bowwow sound}

\vspace{5pt}

\textbf{Example 2:} \\
Sentence: The wind whispers softly through the trees. \\
Span: \texttt{whispers softly}

\vspace{5pt}

\textbf{Example 3:} \\
Sentence: The visual effects in this movie are stunning. \\
Span: \texttt{None}

\vspace{10pt}

\textbf{Input Prompt:} \\
Sentence: \{\{Sentence\}\} \\

Span: \\ 
\bottomrule
\end{tabular}
\caption{Input prompt for generating auditory knowledge span.}
\label{prompt_audio}
\end{table*}

\section{Prompts for span generation}
\label{detail generate prompt}

This section describes the prompt for span labels for auditory knowledge in GLUE. \Cref{prompt_audio} is the prompt for auditory.

\end{document}